\pdfoutput=1

\documentclass[11pt]{article}

\usepackage[final]{acl}

\usepackage{times}
\usepackage{latexsym}

\usepackage[T1]{fontenc}

\usepackage[utf8]{inputenc}

\usepackage{microtype}

\usepackage{inconsolata}

\usepackage{graphicx}

\usepackage{algorithm}
\usepackage{algorithmic}
\usepackage{amsmath}
\usepackage{amssymb}
\usepackage{bm}
\usepackage{booktabs}
\usepackage{multicol}
\usepackage{multirow}
\usepackage{graphicx}
\usepackage{subfigure}
\usepackage{listings}
\usepackage{threeparttable}
\usepackage[utf8]{inputenc}    
\usepackage{listingsutf8}      

\title{Alleviating Distribution Shift in Synthetic Data for Machine Translation Quality Estimation}

\author {
    Xiang Geng\textsuperscript{\rm 1}\footnotemark[1] \quad 
    Zhejian Lai\textsuperscript{\rm 1}\footnotemark[1] \quad
    {\bf Jiajun Chen}\textsuperscript{\rm 1} \quad
    {\bf Hao Yang}\textsuperscript{\rm 2} \quad
    {\bf Shujian Huang}\textsuperscript{\rm 1} \footnotemark[2] \\
    \textsuperscript{\rm 1} National Key Laboratory for Novel Software Technology, Nanjing University, Nanjing, China\\
    \textsuperscript{\rm 2} Huawei Translation Services Center, Beijing, China\\
    \texttt{\{gx,laizj\}@smail.nju.edu.cn} \\
    \texttt{\{yanghao30\}@huawei.com,\{chenjj,huangsj\}@nju.edu.cn}
}

\begin{document}
\maketitle

\renewcommand{\thefootnote}{\fnsymbol{footnote}}
\footnotetext[1]{Equal contribution, in random order.}
\footnotetext[2]{Corresponding author.}
\renewcommand{\thefootnote}{\arabic{footnote}}

\begin{abstract}
Quality Estimation (QE) models evaluate the quality of machine translations without reference translations, serving as the reward models for the translation task.
Due to the data scarcity, synthetic data generation has emerged as a promising solution.
However, synthetic QE data often suffers from distribution shift, which can manifest as discrepancies between pseudo and real translations, or in pseudo labels that do not align with human preferences.
To tackle this issue, we introduce DCSQE, a novel framework for alleviating distribution shift in synthetic QE data.
To reduce the difference between pseudo and real translations, we employ the constrained beam search algorithm and enhance translation diversity through the use of distinct generation models.
DCSQE uses references—i.e., translation supervision signals—to guide both the generation and annotation processes, enhancing the quality of token-level labels.
DCSQE further identifies the shortest phrase covering consecutive error tokens, mimicking human annotation behavior, to assign the final phrase-level labels.
Specially, we underscore that the translation model can not annotate translations of itself accurately.
Extensive experiments demonstrate that DCSQE outperforms SOTA baselines like CometKiwi in both supervised and unsupervised settings.
Further analysis offers insights into synthetic data generation that could benefit reward models for other tasks.
The code is available at \url{https://github.com/NJUNLP/njuqe}.

\end{abstract}

\section{Introduction}
Unlike machine translation (MT) metrics such as BLEU~\cite{BLEU} and Comet~\cite{comet}, which rely on reference translations to evaluate quality, quality estimation (QE) assesses translation quality without any reference~\cite{mtqe}.
QE plays a crucial role in post-editing workflows by reducing human effort through filtering low-quality translations and identifying incorrect segments~\cite{effort}.
From the perspective of large language models (LLMs), QE models can function as reward models~\cite{RLHF} for machine translation.
\citet{QE-as-RM} explores the use of quality estimation to align translation models with human feedback, achieving significant improvements in translation performance.
LLMRefine~\cite{llmrefine} uses LLMs to refine translations according to fine-grained QE feedback.

\begin{figure}[t]
    \centering 
    \includegraphics[width=\columnwidth]{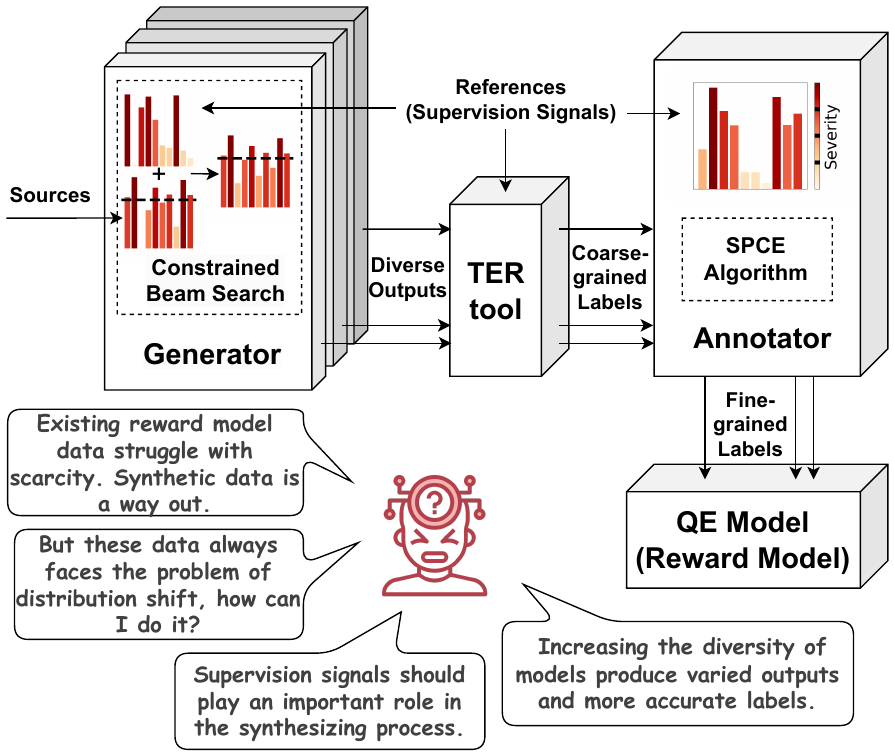} 
    \caption{
    We explore ways to enhance the quality of synthetic QE data by leveraging supervision signals and increasing model diversity.
    The histogram represents the generation probabilities of translation models.
    }
    \label{fig:main}
    \vspace{-15pt}
\end{figure}

The Multidimensional Quality Metrics (MQM)~\cite{MQM_old} annotations have become the primary standard for QE in recent years, as MQM scores are more reliable than the earlier Direct Assessment ~\cite{DA} scores ~\cite{MQM}.
As shown in Table~\ref{tab:example}, the MQM annotations are both fine-grained and explainable, offering not only error spans but also the severity of each error span.
Acquiring these fine-grained MQM annotations is labor-intensive. Consequently, their datasets are typically small and restricted to specific language pairs~\cite{wmt-2023-mt-findings}.

\begin{table}[t]
\centering
\resizebox{\columnwidth}{!}
{
\begin{tabular}{l|l|l}
\hline
\textbf{SRC} & \multicolumn{2}{l}{Echidna with amethyst and magenta spikes .} \\
\textbf{MT} & \multicolumn{2}{l}{\textcolor{red}{Die} Echidna mit \textcolor{red}{Amethyst und Magenta-Spitzen} . } \\
\textbf{REF} & \multicolumn{2}{l}{Das Echidna mit Amethyst- und Magenta-Stacheln .} \\
\hline
\hline
\textbf{ID} & \textbf{Error Span}  & \textbf{Severity} \\
\textbf{No. 1} & \textcolor{red}{Die} & MINOR \\
\textbf{No. 2} & \textcolor{red}{Amethyst und Magenta-Spitzen} & CRITICAL \\
\hline
\hline
\textbf{Labels} & \multicolumn{2}{l}{\textcolor{red}{BAD} OK OK \textcolor{red}{BAD} \textcolor{red}{BAD} \textcolor{red}{BAD} \textcolor{red}{BAD} OK}\\
\textbf{Score} & \multicolumn{2}{l}{-0.375} \\
\hline

\end{tabular}
}
\caption{
At test time, only the source (SRC) and its translation (MT) are available; the reference (REF) is not accessible.
}
\label{tab:example}
\vspace{-15pt}
\end{table}

Therefore, existing studies aim to generate synthetic MQM data for QE from parallel sentences. For instance, MQMQE~\cite{wmt2023} randomly masks spans in the references, replaces these spans with tokens sampled negatively from a translation model, and annotates the severity using the generation probabilities from the model itself.
InstructScore~\cite{instructscore} instructs the GPT-4~\cite{gpt4} to generate errors with specific severities based on references.
 
Although these methods achieve competitive performance, the distribution of their synthetic data may differ significantly from that of real data.
The distribution shift problem not only causes a decrease in QE performance but also in downstream human preference optimization ~\cite{bpo}.
Specifically, the negative sampling strategy renders the synthetic translations of MQMQE less fluent. 
Moreover, the synthetic labels of MQMQE do not align with human preferences, as randomly masked spans often disrupt entire phrases, and the translation model tends to be overly confident in its own outputs.
While InstructScore produces fluent synthetic translations and accurate synthetic labels, the generated errors appear unnatural, unlike those that advanced translation models would typically produce.
More importantly, utilizing powerful closed-source LLMs requires substantial time and financial resources.

In this paper, we propose a framework, DCSQE (as shown in Figure~\ref{fig:main}), \textbf{D}istribution-\textbf{C}ontrolled Data \textbf{S}ynthesis for \textbf{QE}, aiming to alleviate the distribution shift problem when generating synthetic QE data.
Specifically, we first train two translation models as Generator and Annotator respectively.
We then use the Generator to generate synthetic translations with constrained beam search (CBS)~\cite{cbsqe}, which preserves the main structure of references while maximizing generation probabilities. 
This allows us to treat matched tokens as correct with high accuracy using TER~\cite{hter} tool.
For the mismatched part, we utilize generation probabilities provided by the Annotator to rejudge their fine-grained severities.
Since human annotators typically prefer annotating entire phrases as spans, we propose an algorithm, which helps aggregate token-level labels into phrase-level labels.

Extensive experiments across three language directions (English-German, Chinese-English, and Hebrew-English) demonstrate that DCSQE achieves new (SOTA) results in both supervised and unsupervised settings. 
Further analysis offers insights into synthetic data generation that may benefit general reward models: 
(1) distribution shift problem is crucial in synthetic data methods;
(2) diversity between annotation and generation models enhances annotation accuracy;
(3) diversity in generation models provides further improvements;
(4) the capacity of the generation model must be balanced;
(5) enhancing the capacity of the annotation model with supervision signals is helpful.

\section{Background}

\textbf{Quality estimation.}
The quality estimation task assesses the quality of translation without access to references.
Given a source sentence $\mathbf{x}$ and its machine translation $\hat{\mathbf{y}}= \{y_1, y_2, \dots, y_n\}$ with $n$ words, we aim to predict the following quality labels: 
(1) Span-level MQM labels $\mathbf{h}=\{h_1, h_2, \dots, h_n\}$, the label $h_i$ is categorically annotated with error severity (MINOR, MAJOR, or CRITICAL) through professional human annotation. 
(2) Word-level labels $\mathbf{g}=\{g_1, g_2, \dots, g_n\}$, the label $g_i$ is usually a binary label (OK or BAD), which is derived from the span-level label. 
Words within error spans are marked as ``BAD'', and vice versa. 
(3) Sentence-level MQM scores $s$, which is derived from the span-level labels. 
The scores can be calculated as 
\begin{align}\label{eq:mqm}
    s = 1 - \frac{n_{\text{MINOR}}+5\times n_{\text{MAJOR}}+10\times n_{\text{CRITICAL}}}{n},
\end{align}
where $n_{\text{severity}}$ denotes the number of each error severity and $n$ denotes the translation length.

\textbf{Model architecture.} Previous works, e.g. CometKiwi~\cite{CometKiwi2023} and MQMQE~\cite{wmt2023}, usually adopt multilingual pre-training language model, typically XLM-R~\cite{xlmr}, as the backbone architecture for the QE model.
The outputs from the final layer of the model are used to derive representations for each token.
Word-level representations are obtained by averaging the representations of all tokens within a word. 
Similarly, the regression score representation is computed by averaging the representations of all target tokens. 
These representations are subsequently fed into linear layers to predict word-level quality labels and regression scores, respectively.

\textbf{Training.} As noted in~\cite{wmt2023}, the span-level task can be regarded as a word-level task.
Thus, the overall objective only combines the sentence-level and word-level tasks.
The sentence-level task is treated as a regression problem optimized with MSE loss, while the word-level task is framed as a sequence labeling problem using cross-entropy loss.
In a supervised setting, the model is firstly pre-trained on synthetic data and then fine-tuned on real data; while unsupervised setting only trained using synthetic data.

\textbf{Inference.} The error severity can be determined by comparing the probability of ``OK'' against various thresholds.
For the span-level task, consecutive ``BAD'' tokens are identified as a span, with their error severity determined by the worst error severity within the span.

\section{Method}

In this section, we first describe the process of generating realistic synthetic translations. Following this, we introduce methods for annotating synthetic labels to ensure they accurately align with human preferences.

\subsection{Synthetic Translations}

To produce synthetic translations that closely resemble real machine translations, we directly generate synthetic translations using the translation model with the constrained beam search (CBS) algorithm~\cite{cbsqe}.
Similar to the standard beam search (BS) algorithm, the CBS algorithm also seeks to generate translations with high generation probabilities, thereby producing natural translation errors.
However, the standard BS algorithm typically generates synonyms of the references, making it difficult to obtain accurate synthetic labels. 
In contrast, the CBS algorithm is designed to preserve reference tokens when their generation probabilities exceed a specified threshold, reducing the risk of generating synonyms.

In most QE applications, the target translation model is not accessible. 
As a result, existing studies train a surrogate translation model, referred to as the Generator, training on parallel corpora.
We assume that increasing the diversity of synthetic translations enhances the likelihood of generating translations that are more similar to the target.
However, as a variation of the BS algorithm, the CBS algorithm also struggles to generate diverse translations.
In preliminary experiments, we attempt to perturb Generators using dropout~\cite{dropout}, as suggested by~\cite{unsupervised_qe}, to produce diverse translations for a single source sentence. 
However this approach does not enhance the diversity of synthetic translations, nor does it improve QE performance.
Therefore, we further explore enhancing the diversity of Generators by training them on different parallel subsets in Section~\ref{sec:diversity_analysis}.

\begin{figure*}[ht]
    \centering
    \includegraphics[width=\textwidth]{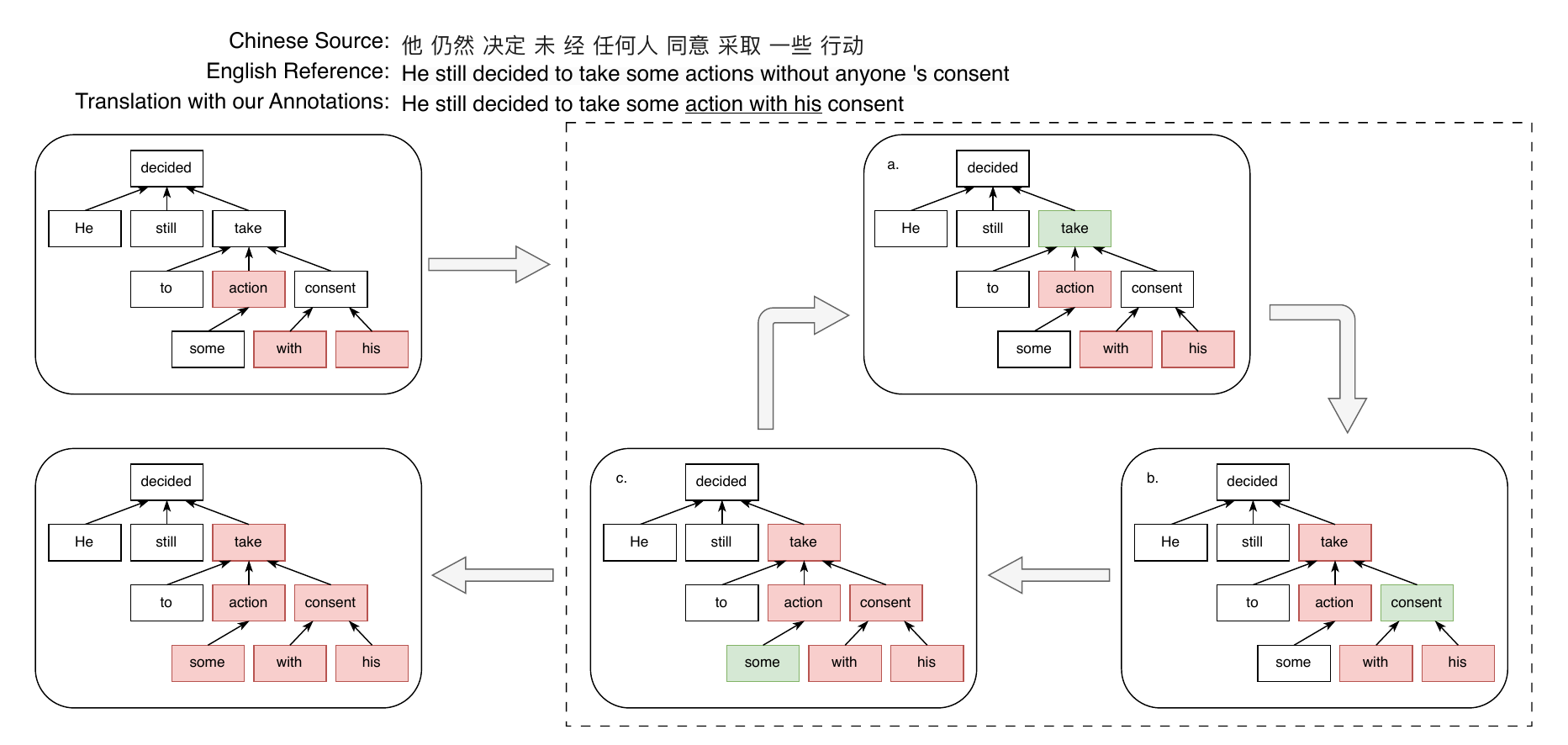} 
    \caption{The illustration of the SPCE algorithm. 
    }
    \label{fig:SPCE}
    \vspace{-10pt}
\end{figure*}

\subsection{Synthetic Labels}

\textbf{Coarse-grained labels.}
CBS allows us to preserve the main structure of references, thereby obtaining accurate labels using the exact match.
We use the TER tool to perform the word-level alignment between synthetic translation and the corresponding reference.
The match part is regarded as “OK”, and vice versa.
After generating synthetic translations, we need to annotate them to align with human preference.

\textbf{Refined labels.}
To assign fine-grained severities to each ``BAD'' token and correct false-negative labels, we use a translation model as the Annotator to rejudge the previous labels.
As demonstrated in~\cite{unsupervised_qe} and~\cite{ssqe}, the confidence of the translation model, i.e. the generation probabilities, serves as a reliable indicator for assessing translation quality.
Therefore, we utilize the generation probability $p_i$ of each token to determine its error severity.
Specifically, the severity label $\hat{h_i}$ is assigned as follows:
\begin{align}
\label{eq:get_severity}
\hat{h}_i&=\begin{cases}
\text{CRITICAL} &0 \le p_i < t_{\text{CRITICAL}} \\
\text{MAJOR} & t_{\text{CRITICAL}} \le p_i < t_{\text{MAJOR}} \\
\text{MINOR} & t_{\text{MAJOR}} \le p_i < t_{\text{MINOR}} \\
\text{OK} & {t_{\text{MINOR}} \le p_i \le 1} \\
\end{cases},
\end{align}
where $t_{\text{MINOR}} < t_{\text{MAJOR}} <t_{\text{CRITICAL}}$ are three ordered thresholds that can be determined using a validation dataset.

\label{sec:syn_labels}
\textbf{Leveraging supervision signals.}
When the model lacks the necessary translation knowledge for the given input, its generation probabilities become speculative and fail to meaningfully correlate with the severity of errors.
To address this issue, we ensure that the Annotator is trained on parallel pairs that are also used for generating synthetic data.
Therefore, the Annotator becomes ``professional'' when working with the given parallel pairs, rather than ``amateur''.

\textbf{Differentiate the Annotator from the Generator.}
MQMQE~\cite{wmt2023} utilizes the same translation model to annotate its own outputs. However, this approach may lead to overconfidence in the model's predictions, resulting in inaccurate annotations.
To investigate this problem, we introduce another different, well-trained translation model to serve as the Annotator.

\textbf{Combine error tokens into human-readable phrases.}
Although we possess token-level labels, human annotators tend to prefer annotating entire phrases that contain translation errors to ensure the clarity and interpretability of the annotations.
At the same time, annotators strive to make the phrase as short as possible, ensuring it covers all consecutive error tokens without including any unnecessary ones.
Therefore, we propose the Shortest Phrase Covering Errors (SPCE) algorithm to combine error tokens into human-readable phrases.
As illustrated in Figure~\ref{fig:SPCE}, the SPCE algorithm operates as follows:

First, we parse the synthetic translation to obtain its dependency tree. 
Consecutive ``BAD'' tokens are added to the candidate set, which serves as the initial phrase.
Following this, we iteratively execute the following steps:
\begin{itemize}
    \item Step a. We employ the Lowest Common Ancestor (LCA) algorithm~\cite{lca} to find the common ancestor node of the candidate tokens (i.e. word ``take''). 
    This step locates the smallest sub-tree that encompasses all errors.
    \vspace{-5pt}
    \item Step b. To ensure syntactic coherence, we add all tokens along the path from the LCA into the candidate set (i.e. word ``consent'' which is on the path from ``his'' to ``take'').
    \vspace{-5pt}
    \item Step c. To ensure the candidate tokens are consecutive, we add all tokens, which are located between the leftmost and rightmost candidate tokens, into the candidate set (i.e. word ``some'' which is in the phrase ``take some action with his consent'').
\end{itemize}
The iteration terminates until no additional tokens need to be added to the candidate set.
We provide the detailed algorithm in Appendix~\ref{sec:spce_details}.

To assign a representative fine-grained label to each phrase, we determine its severity by selecting the most severe error label among the candidate tokens.

\section{Experiment}
\label{sec:exp}

In the experiment section, we aim to address the following research questions regarding DCSQE:
(1) How does DCSQE perform across supervised and unsupervised settings for various language pairs?
(2) To align synthetic labels with human preferences, we introduce the Annotator and the SPCE algorithm to generate  MQM data. To what extent are these techniques effective?
(3) Can the Annotator and the Generator be the same model? In other words, can a model fairly annotate its own output?
(4) Does increasing the diversity of Generators lead to enhanced QE performance?
(5) How does the translation performance of the Generator and the Annotator influence the quality of the synthetic data?
(6) Does DCSQE demonstrate advantages in terms of generation cost and convergence efficiency?

To address \textbf{Q1}, we evaluate DCSQE in both supervised and unsupervised settings across multiple language pairs (EN-DE, ZH-EN, HE-EN).
For \textbf{Q2}, we conduct ablation studies to quantify the individual contributions of the Annotator and the SPCE algorithm in improving the alignment of synthetic labels with human preferences.
Regarding \textbf{Q3}, we investigate the feasibility of employing a single model to simultaneously serve as both the Generator and the Annotator.
To explore \textbf{Q4}, we enhance the diversity of Generators by training them on distinct parallel subsets.
To address \textbf{Q5}, we control the translation performance of the Generator and the Annotator by training them on corpora of varying sizes.
Finally, for \textbf{Q6}, we compare DCSQE against other synthetic data approaches in terms of generation cost.

\subsection{Experiment Setup}
\paragraph{Datasets.}

In our experiment, we utilized the dataset provided by the Workshop on Machine Translation (WMT) QE Shared Task~\cite{wmt2023-findings}. 
The dataset comprises two types of data: \textbf{parallel data}\footnote{\url{https://www2.statmt.org/wmt23/translation-task.html\#training}} and \textbf{MQM data}\footnote{\url{https://wmt-qe-task.github.io/wmt-qe-2023/subtasks/resources}}.
Data statistics are given in Appendix~\ref{sec:data_statics}.

We employed parallel datasets for three language pairs: English-German (EN-DE), Chinese-English (ZH-EN), and Hebrew-English (HE-EN). 
Parallel data is widely used in the QE community including data synthesis, enhancement of translation knowledge, etc.
Unless specified, the parallel sentences employed for generating synthetic data and the training set for the Annotator are kept disjoint and the parallel sentences used for generation are derived from a subset of the Annotator's training set.

We utilize the MQM training set from WMT2023 for EN-DE and ZH-EN language pairs.
For evaluation, we employ the MQM test set from WMT2022, which includes EN-DE and ZH-EN; the MQM test set from WMT2023, which includes EN-DE, ZH-EN, and HE-EN.
We exclude the WMT2022 test set from the supervised setting due to its overlap with the WMT2023 training set.
Furthermore, span-level evaluation is not performed on the WMT2022 test set, as it lacks the necessary span-level annotations.

\paragraph{Baselines.}
We incorporated top-performing baselines in our experiments:
\textbf{CometKiwi}~\cite{CometKiwi2023} stands out as the SOTA QE model, widely adopted in translation studies~\cite{wmt-2023-mt-findings}.
CometKiwi enhances its generalization capabilities by leveraging labeled QE datasets with various annotations\footnote{\url{https://github.com/sheffieldnlp/mlqe-pe}}, across multiple language pairs. 
Additionally, CometKiwi is built on the XLMR-XL model~\cite{xlmr}, which has seven times more parameters than ours.
\textbf{GEMBA-MQM}~\cite{gemba-mqm} employs few-shot prompts to guide GPT-4 in generating MQM predictions.
\textbf{InstructScore}~\cite{instructscore} and \textbf{MQMQE}~\cite{wmt2023} are two representative synthetic data approaches.
InstructScore generates MQM data by prompting GPT-4, whereas MQMQE employs a translation model combined with negative sampling to produce synthetic data. 

\begin{table*}[t]
\centering
\footnotesize
\begin{tabular}{ccccccccccc}
\toprule
\multirow{2}{*}{\textbf{Setting}} & \multirow{2}{*}{\textbf{Dataset}} & \multirow{2}{*}{\textbf{Method}} & \multicolumn{2}{c}{\textbf{Sentence-level}} & \multicolumn{2}{c}{\textbf{Word-level}} & \multicolumn{3}{c}{\textbf{Span-level}} \\
\cmidrule(lr){4-5} \cmidrule(lr){6-7} \cmidrule(lr){8-10}
& & & \textbf{Spearman} & \textbf{Pearson} & \textbf{MCC} & \textbf{F1} & \textbf{F1} & \textbf{Prec} & \textbf{Recall} \\
\midrule
\multirow{10}{*}{{\rotatebox{90}{Supervised}}} & \multirow{5}{*}{23 EN-DE} & CometKiwi\textsuperscript{\dag} & 40.47 & 40.97 & 21.50 & - & 23.50 & - & - \\
 & & GEMBA & 40.06 & 35.12 & 13.21 & 18.17 & \hspace{5pt}5.40 & \hspace{5pt}7.50 & \hspace{5pt}4.22 \\
& & InstructScore & 35.03 & 32.85 & 22.54 & 26.60 & 23.77 & 18.19 & 34.46 \\
& & MQMQE & 37.88 & 25.07 & 22.84 & 25.22 & 21.14 & 17.13 & 27.62 \\
& & DCSQE & \textbf{43.17} & 41.64 & \textbf{27.11} & 30.61 & \textbf{25.89} & 21.20 & 33.26 \\
\cmidrule(lr){2-10}
&\multirow{5}{*}{23 ZH-EN} & CometKiwi\textsuperscript{\dag} & 40.35 & 35.53 & 26.90 & - & 27.20 & - & - \\
& & GEMBA & 33.80 & 32.56 & 16.11 & 18.21 & \hspace{5pt}9.23 & \hspace{5pt}8.41 & 10.22 \\
& & InstructScore & 36.40 & 28.00 & 26.05 & 29.54  & 26.56 & 23.87 & 29.94 \\
& & MQMQE & 39.26 & 22.94 & 23.53 & 26.49 & 22.01 & 22.65 & 21.40 \\
& & DCSQE & \textbf{46.41} & 37.55 & \textbf{28.12} & 28.61 & \textbf{27.71} & 22.19 & 36.88 \\
\midrule
\multirow{17}{*}{{\rotatebox{90}{Unsupervised}}} & \multirow{5}{*}{23 HE-EN} & CometKiwi\textsuperscript{\dag} & 55.00 & 44.15 & 33.40 & - & 10.50 & - & - \\
& & GEMBA & 54.63 & 35.27 & 21.79 & 28.39 & 12.12 & 14.87 & 10.23 \\
& & InstructScore & 31.72 & 33.18 & 32.39 & 37.25 & 36.29 & 30.84 & 44.08 \\
& & MQMQE & 25.90 & 15.66 & 16.19 & \hspace{5pt}8.39 & 23.78 & 23.57 & 23.99 \\
& & DCSQE & \textbf{56.46} & 45.06 & \textbf{36.34} & 38.28 & \textbf{39.51} & 42.25 & 37.10 \\
\cmidrule(lr){2-10}
& \multirow{4}{*}{23 EN-DE} & InstructScore & 12.08 & 20.16 & \textbf{18.97} & 22.59 & 19.70 & 14.94 & 28.95 \\
& & MQMQE & 24.11 & 20.99 & \hspace{5pt}7.49 & \hspace{5pt}2.39 & 16.79 & 19.80 & 14.58  \\
& & DCSQE & \textbf{35.78} & 37.19 & 18.00 & 21.91 & \textbf{20.15} & 16.27 & 26.46 \\
\cmidrule(lr){2-10}
& \multirow{3}{*}{23 ZH-EN} & InstructScore & 30.46 & 30.11 & 21.71 & 23.88 & 21.60 & 14.93 & 39.34 \\
& & MQMQE & \hspace{5pt}6.52 & 19.35 & \hspace{5pt}3.07 & \hspace{5pt}1.13 & 14.05 & 19.80 & 10.89 \\
& & DCSQE & \textbf{37.54} & 28.04 & \textbf{23.41} & 26.45 & \textbf{23.67} & 20.52 & 27.98 \\
\cmidrule(lr){2-10}
& \multirow{4}{*}{22 EN-DE} & InstructScore & 24.00 & 35.07 & 22.32 & 22.05 & - & - & - \\
& & MQMQE & 40.22 & 36.15 & 12.07 & \hspace{5pt}4.92 & - & - & - \\
& & DCSQE & \textbf{41.27} & 40.88 & \textbf{24.36} & 25.44 & - & - \\
\cmidrule(lr){2-10}
& \multirow{3}{*}{22 ZH-EN} & InstructScore & 13.53 & 25.75 & \hspace{5pt}5.53 & \hspace{5pt}6.58 & - & - & - \\
& & MQMQE & \hspace{1pt}-0.88 & 33.99 & -0.35 & \hspace{5pt}0.00 & - & - & - \\
& & DCSQE & \textbf{26.27} & 44.04 & \textbf{10.27} & 11.57 & - &  - & - \\
\bottomrule
\end{tabular}
\caption{Main results on different QE test sets. We follow the setting in GEMBA-MQM~\cite{gemba-mqm}, which utilizes a few-shot prompt containing examples for EN-DE and ZH-EN language pairs. 
However, since the prompt does not include examples for HE-EN, GEMBA-MQM is treated as a supervised method for EN-DE and ZH-EN, while an unsupervised one for HE-EN.
\textsuperscript{\dag} The details of CometKiwi are described in the second paragraph of Section~\ref{sec:implementation_detail}.
}
\vspace{-10pt}
\label{tab:main_result}
\end{table*}

\paragraph{Implementation Details.}
\label{sec:implementation_detail}
The Generator and the Annotator for synthesizing pseudo MQM data are based on the Transformer-Large~\cite{transformer} architecture with a shared decoder input-output embedding.
We use the TER tool called TERCOM\footnote{\url{https://www.cs.umd.edu/~snover/tercom/}} to annotate the synthetic translations generated by the Generator.
We train the QE model using the XLMR-L backbone~\cite{xlmr} for all synthetic data approaches. 
The experiments are implemented using the open-source Fairseq toolkit~\cite{fairseq} and conducted on NVIDIA V100 GPUs.

Since CometKiwi does not provide sentence-level results for the individual model, we reproduce sentence-level results using the released \textbf{CometKiwi-XL}\footnote{\url{https://huggingface.co/Unbabel/wmt23-CometKiwi-da-xl}}. 
For word- and span-level results, we directly utilize the results provided in the CometKiwi report~\cite{CometKiwi2023} (where model size is greater than or equal to \textbf{XL}), as the corresponding word- and span-level models have not been released.
InstructScore is implemented using 10K synthetic data released by~\citet{instructscore}. 
For DCSQE and MQMQE, we generate 500K synthetic data in each language pair.
In Section~\ref{sec:efficiency}, we demonstrate that DCSQE remains competitive with InstructScore when trained on 10K synthetic data while requiring significantly lower costs.
Additional implementation details are provided in Appendix~\ref{sec:appendix oid}.

\paragraph{Evaluations.}
Following WMT23 QE shared tasks~\cite{wmt2023-findings}, sentence-level evaluation utilizes Spearman's rank correlation coefficient as the primary metric, complemented by Pearson's correlation coefficient.
For word-level evaluation, the Matthews Correlation Coefficient (MCC) serves as the primary metric, complemented by F1 score.  
For span-level evaluation, the primary metric is the weighted F1 score, which accounts for all error severities.
We mark the results with \textbf{bold} if the results are statistically significant ($p<0.05$) under Williams significance test~\cite{williams_test}.

\subsection{Main Results}
\paragraph{Supervised setting.}
As demonstrated in Table~\ref{tab:main_result}, DCSQE substantially outperforms CometKiwi despite utilizing fewer parameters, achieving notable improvements with average gains of 4.38 in Spearman, 3.41 in MCC, and 1.45 in F1 score.
Moreover, DCSQE significantly outperforms GEMBA-MQM, which is based on the advanced LLM, GPT-4.
Compared to other synthetic methods, i.e., MQMQE and InstructScore, DCSQE demonstrates consistent superiority, indicating that our synthetic data is of higher quality.

\paragraph{Unsupervised Setting.}
As shown in Table~\ref{tab:main_result}, both MQMQE and InstructScore demonstrate significant performance declines compared to their supervised counterparts, with reductions of 15.74 and 7.64 on average, respectively.
That implies the distribution shift problem in previous synthetic QE data. 
In contrast, our proposed method achieves superior robustness, incurring a smaller average reduction of 6.64 points. 
Moreover, DCSQE outperforms CometKiwi, which relies on labeled datasets from other language pairs, on HE-EN. This indicates that our synthetic data provides more relevant QE knowledge for HE-EN compared to datasets from different language pairs.

\subsection{Ablation Study}

The DCSQE framework introduces two techniques to align the synthetic labels with human preferences: 
(1) leveraging the Annotator to rejudge tokens initially labeled as BAD by the TER tool, 
and (2) applying the SPCE Algorithm to aggregate token-level annotations into span-level annotations.
The effectiveness of these techniques within the DCSQE framework is empirically validated in Table~\ref{tab:ablation_study}. 
The results demonstrate that the Annotator effectively corrects errors in coarse-grained labels.
Furthermore, the SPCE algorithm successfully identifies phrase spans, thereby achieving better alignment with human preferences.

\begin{table}[t]
\centering
\footnotesize
\begin{tabular}{lcc}
\toprule
\textbf{Method} & \textbf{Spearman}$\uparrow$ & \textbf{MCC}$\uparrow$ \\
\midrule
DCSQE & 35.78 & 18.00 \\
\hspace{5pt} -SPCE & 30.99 & 15.70 \\
\hspace{5pt} -SPCE \& Annotator & 11.24 & 11.17 \\
\bottomrule
\end{tabular}
\caption{Ablation studies on the WMT23 EN-DE test set.}
\label{tab:ablation_study}
\end{table}

\subsection{Analysis}

In this subsection, we aim to investigate the contributions of the Generator and the Annotator to the performance of DCSQE. 
We create translation models with varying levels of performance by training them on datasets of different sizes.
We train three translation models—$S$, $M$, and $L$—using 1M, 5M, and 20M distinct sentence pairs, respectively. 
To enhance diversity while maintaining similar performance, we then train three additional models—$S'$, $M'$, and $L'$—on the remaining 1M, 5M, and 20M sentence pairs.
All subsequent analyses are conducted on the WMT23 EN-DE test set in the unsupervised setting.

\begin{table}[t]
\centering
\resizebox{\columnwidth}{!}{\begin{tabular}{ccccc}
\toprule
\textbf{Generator} & \textbf{Annotator} & \textbf{Error Rate (\%)} & \textbf{Spearman}$\uparrow$ & \textbf{MCC}$\uparrow$ \\
\midrule
$M$& $M$ & \hspace{5pt}1.60 & 25.91 & 10.36 \\
$L$& $L$& \hspace{5pt}0.11 & 29.00 & 11.61 \\
$M$& $L$ & 19.23 & 35.78 & 18.00 \\
\bottomrule
\end{tabular}}
\caption{Analysis comparing individual and collaborative results of deploying models $L$ and $M$.}
\label{tab:calibrate_study}
\vspace{-10pt}
\end{table}

\textbf{The model cannot fairly annotate its outputs.}
\label{sec:annotate_itself}
As discussed in Section~\ref{sec:syn_labels}, the translation model may exhibit overconfidence in its output, leading to an increased number of ``OK'' labels.
To investigate that, we calculate the error rate across different settings.
Table~\ref{tab:calibrate_study} shows that translation models tend to consistently assign more ``OK'' labels to their own output, regardless of the model's performance.
The poor QE performance implies that most of the ``OK'' labels are false-negative.
We also provide a corresponding case study (Table~\ref{tab:annotate_itself_case_study} see in Appendix).

\textbf{Diversity of Generators enhances QE performance.}
\label{sec:diversity_analysis}
To investigate the impact of the Generator's diversity on the QE performance, we employ $L$ and $L'$ to generate synthetic data.
We quantify the diversity between $L$ and $L'$ by calculating the BLEU score between their output for the same source.
The average BLEU score on Flores-200\footnote{\url{https://github.com/facebookresearch/flores/blob/main/flores200/README.md}} is 80.06.
The diversity result in distinct translation errors, which are more likely to comprehensively cover realistic errors.
As a result, as shown in Table~\ref{tab:generator_diversity}, generating diverse synthetic data for the identical parallel pair also enhances the QE performance.

\begin{table}[t]
    \centering
    \footnotesize
    \begin{tabular}{c c c c}
        \toprule
        \textbf{Generator} & \textbf{Annotator} & \textbf{Spearman}$\uparrow$ & \textbf{MCC}$\uparrow$ \\
        \midrule
        $L$ & $M$ & 31.19 & 10.08 \\ 
        $L+L'$ & $M$ & 32.19 & 11.05 \\
        \bottomrule
    \end{tabular}
    \caption{The impact of diversity in Generators.}
    \label{tab:generator_diversity}
\end{table}

\textbf{The capacity of the Generators must be balanced.}
If the Generator is too strong, the generated data will have few errors, limiting learning opportunities; if the Generator is too weak, the data will be too noisy or unrealistic.
To investigate this point, we measure the \textit{Error Rate} of synthetic translations and the \textit{Similarity} between synthetic and real translations across different Generators in Table~\ref{tab:generator_info}.
To be specific, the \textit{Similarity} is measured by the BLEU score, which calculates between synthetic translations and the ``real'' translations from the WMT2023 QE validation set.

\begin{table}[t]
    \centering
    \footnotesize
    \begin{tabular}{c c c}
        \toprule
        \textbf{Generator} & \textbf{Error Rate (\%)} & \textbf{BLEU} \\
        \midrule
        $S$ & 32.68 & 31.02 \\ 
        $M$ & 27.02 & 43.08 \\ 
        $L$ & \hspace{5pt}0.58 & 51.13 \\
        \bottomrule
    \end{tabular}
    \caption{Metrics on Generators.}
    \label{tab:generator_info}
    \vspace{-10pt}
\end{table}

\begin{figure*}[htbp]
    \centering
    \includegraphics[width=\textwidth]{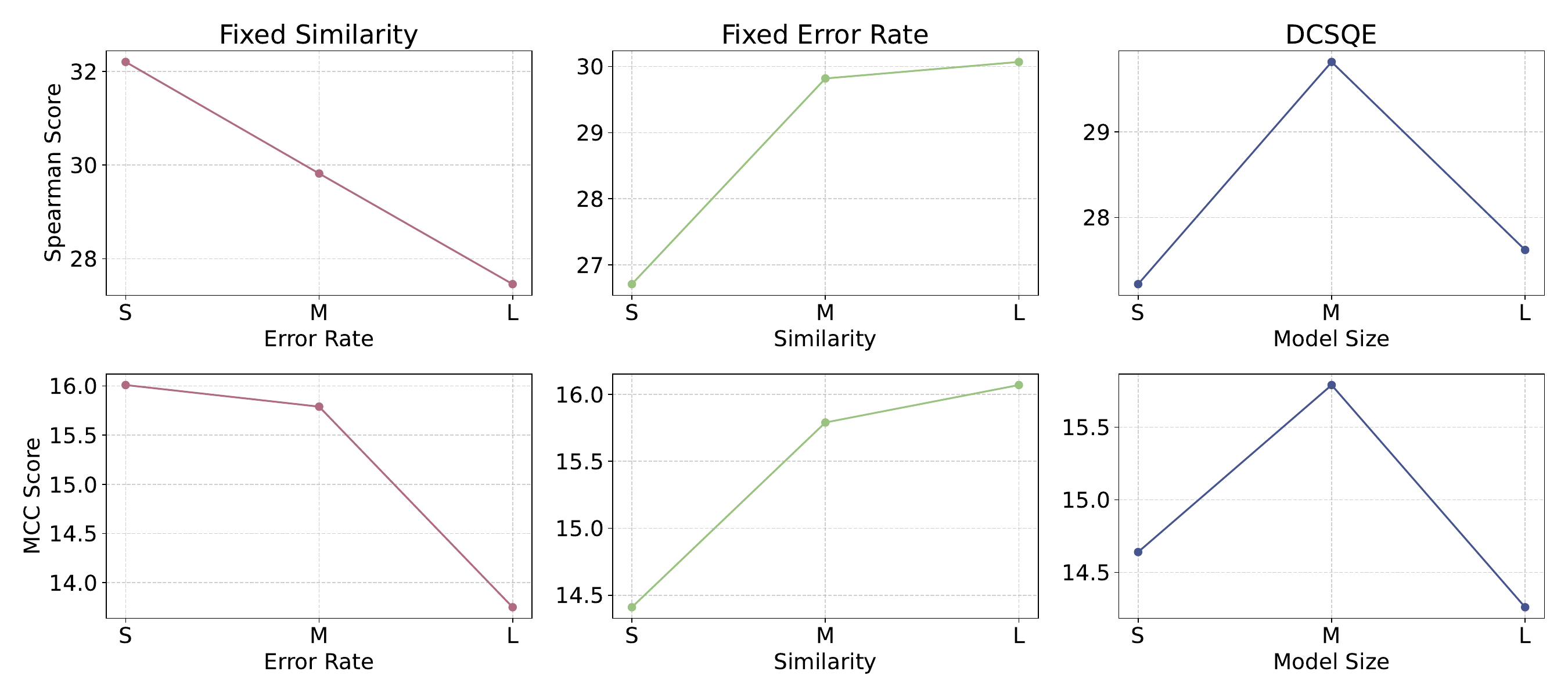}
    \caption{Impact of Generator metrics on synthetic data quality in sentence-level and word-level tasks. The first row represents the sentence-level results. The second row represents the word-level results.} 
    \label{fig:generator_analysis} 
\end{figure*}
To isolate the individual impact of each factor on QE performance , we employ the following downsampling strategy to select subsets of translations from one Generator—referred to as the original Generator—that match the \textit{Error Rate} distribution of another Generator.
Specifically, we begin by sampling a set of \textit{Error Rate} scores from the \textit{Error Rate} distribution of one Generator. 
For each sampled \textit{Error Rate} score, we select a translation, whose \textit{Error Rate} score is closest to the sampled score, from the original Generator.
All samples are drawn without replacement.
This downsampled subset retains the translation style (\textit{Similarity}) of the original Generator while mirroring the \textit{Error Rate} distribution of another Generator.

We consider three experimental settings. In the ``Fixed Similarity'', we fix the \textit{Similarity} by exclusively deploying model M as the original Generator and downsample three subsets to mirror the \textit{Error Rate} distribution of S, M, and L. 
Conversely, in the ``Fixed Error Rate'', we exclusively mirror the \textit{Error Rate} distribution of M and downsample data from original Generators S, M, and L to diversify the \textit{Similarity} level.
We also include the ``DCSQE'' for comparison.
As shown in Figure~\ref{fig:generator_analysis}, when the \textit{Similarity} is fixed, the performance degradation as the \textit{Error Rate} decreases. 
Similarly, when the \textit{Error Rate} is fixed, the performance improves with increasing \textit{Similarity}. 
That explains why DCSQE achieves the highest performance using Generator $M$.

\textbf{Enhancing the capacity of annotation model with supervision signals is helpful.}
\label{sec:enhance_annotator}
There are two possible solutions to enrich the Annotator's translation knowledge: (1) leveraging the parallel data from the Annotator's training set for synthetic data generation, and (2) expansion of the training corpus.
To evaluate their effectiveness, we conduct a comparative analysis of performance across different Annotators in Table~\ref{tab:annotator_study}.
The $M_\text{amateur}$ denotes the Annotator is not trained on the parallel pair for the generation.
The results demonstrates that both solutions enhance the capacity of the annotation model resulting in better QE performance.
The case study (Table~\ref{tab:annotator_case_study} see in Appendix) demonstrates a similar conclusion.

\begin{table}[t]
\centering
\footnotesize
\begin{tabular}{cccc}
\toprule
 \textbf{Generator} & \textbf{Annotator} & \textbf{Spearman}$\uparrow$ & \textbf{MCC}$\uparrow$ \\
\midrule
$M$ & $M_\text{amateur}$ & 29.12 & 13.84 \\
$M$ & $M'$ & 30.43 & 15.01 \\
$M$ & $L$ & 35.78 & 18.00 \\
\bottomrule
\end{tabular}
\caption{Comparison among different Annotators.}
\label{tab:annotator_study}
\vspace{-10pt}
\end{table}

\subsection{Generation Cost}
\label{sec:efficiency}
We measure the generation cost for various synthetic data approaches using 10K samples.
As shown in Table~\ref{tab:generation_time}, DCSQE generates synthetic data 14.29$\times$ faster than InstructScore.
Meanwhile, with the same amount of synthetic data, DCSQE also achieves a notable improvement of 14.29 in the Spearman score with slight decline of 1.76 in MCC.
Despite introducing rejudge and SPCE algorithm, DCSQE does not substantially increase computational complexity, requiring only 3.12$\times$ the timelapse of MQMQE.
Further analysis of DCSQE time overhead is provided in Section~\ref{sec:time_analysis}.

\begin{table}[h]
\centering
\footnotesize
\resizebox{\columnwidth}{!}{
\begin{tabular}{c c c}
\toprule
\textbf{Method} &  \textbf{Generation Time (ms)} $\downarrow$ & \textbf{Speed} $\uparrow$ \\
\midrule
MQMQE & \hspace{10pt}32.76 & $\text{44.65}\times$ \\
InstructScore & 1462.64 & $\text{1}\times$ \\
DCSQE & \hspace{5pt}102.36 & $\text{14.29}\times$ \\
\bottomrule
\end{tabular}}
\caption{
Average generation time per sample of different synthetic data methods on a single V100 GPU.
}
\vspace{-10pt}
\label{tab:generation_time}
\end{table}

\section{Related Works}
Early QE approaches predominantly relied on handcrafted features (e.g. alignment-based confidence~\cite{quest}, source complexity~\cite{scarton2015ushef,shah2015investigating}) designed to capture linguistic and statistical indicators.
Recently, \citet{unsupervised} and \citet{ssqe} regard generation probability of neural models as QE features.

CometKiwi~\cite{CometKiwi2023} aims to transfer QE knowledge by leveraging labeled datasets across diverse annotations and languages.
However, this idea still faces the pitfall of distribution shift, as highlighted in \cite{zouhar2024pitfalls}, primarily due to the scarcity of QE data.
To address this limitation, synthetic data generation methods have emerged.
(1) Negative sampling-based approaches like DirectQE~\cite{directqe} and MQMQE~\cite{wmt2023}, utilize translation models to generate synthetic errors through negative sampling, then directly regard error rate as sentence-level scores; (2) BSQE~\cite{nmt_ter} and CBSQE~\cite{cbsqe}, which generate synthetic translations based on search algorithms and derive labels by matching synthetic translations with references.
In this work, we aim to alleviate the distribution shift problem by leveraging supervision signals and
increasing model diversity.

\section{Conclusion}
The distribution shift of synthetic data poses a persistent challenge in the QE field.
To address this challenge, we propose the DCSQE framework, which mitigates distribution shift by leveraging translation references, a form of translation supervision signal, to guide the generation of both diverse synthetic translations and their corresponding synthetic labels.
Experiments show that DCSQE achieves SOTA results in both supervised and unsupervised settings. 
Furthermore, our analysis underscores some insights for synthetic data generation, which could benefit synthetic data methods for general reward models.

\section*{Limitations}
Our framework is subject to several limitations. 
Firstly, while synthetic data quality correlates with the translation performance of the Annotator, where higher-performance synthesized data is more aligned with human preferences, we did not explore LLMs as Annotators due to computational constraints. 
Secondly, although our method proves effective in high-resource settings, its robustness in extreme data scarcity scenarios (i.e. even parallel datasets are unavailable) needs further validation. 
Thirdly, our insights for synthetic QE data could be applied to general reward models, which need further exploration.

\section*{Acknowledgments}
We would like to thank the anonymous reviewers for their insightful comments. 
Shujian Huang is the corresponding author. 
This work is supported by National Science Foundation of China (No. 62376116, 62176120), the Fundamental Research Funds for the Central Universities (No. 2024300507).

\bibliography{acl2025}

\begin{thebibliography}{39}
\providecommand{\natexlab}[1]{#1}

\bibitem[{Aho et~al.(1973)Aho, Hopcroft, and Ullman}]{lca}
Alfred~V Aho, John~E Hopcroft, and Jeffrey~D Ullman. 1973.
\newblock On finding lowest common ancestors in trees.
\newblock In \emph{Proceedings of the fifth annual ACM symposium on Theory of computing}, pages 253--265.

\bibitem[{Blain et~al.(2023)Blain, Zerva, Rei, Guerreiro, Kanojia, C.~de Souza, Silva, Vaz, Jingxuan, Azadi, Orasan, and Martins}]{wmt2023-findings}
Frederic Blain, Chrysoula Zerva, Ricardo Rei, Nuno~M. Guerreiro, Diptesh Kanojia, Jos{\'e}~G. C.~de Souza, Beatriz Silva, T{\^a}nia Vaz, Yan Jingxuan, Fatemeh Azadi, Constantin Orasan, and Andr{\'e} Martins. 2023.
\newblock \href {https://doi.org/10.18653/v1/2023.wmt-1.52} {Findings of the {WMT} 2023 shared task on quality estimation}.
\newblock In \emph{Proceedings of the Eighth Conference on Machine Translation}, pages 629--653, Singapore. Association for Computational Linguistics.

\bibitem[{Conneau et~al.(2020)Conneau, Khandelwal, Goyal, Chaudhary, Wenzek, Guzm{\'a}n, Grave, Ott, Zettlemoyer, and Stoyanov}]{xlmr}
Alexis Conneau, Kartikay Khandelwal, Naman Goyal, Vishrav Chaudhary, Guillaume Wenzek, Francisco Guzm{\'a}n, Edouard Grave, Myle Ott, Luke Zettlemoyer, and Veselin Stoyanov. 2020.
\newblock \href {https://doi.org/10.18653/v1/2020.acl-main.747} {Unsupervised cross-lingual representation learning at scale}.
\newblock In \emph{Proceedings of the 58th Annual Meeting of the Association for Computational Linguistics}, pages 8440--8451, Online. Association for Computational Linguistics.

\bibitem[{Cui et~al.(2021)Cui, Huang, Li, Geng, Zheng, Huang, and Chen}]{directqe}
Qu~Cui, Shujian Huang, Jiahuan Li, Xiang Geng, Zaixiang Zheng, Guoping Huang, and Jiajun Chen. 2021.
\newblock Directqe: Direct pretraining for machine translation quality estimation.
\newblock In \emph{Proceedings of the AAAI Conference on Artificial Intelligence}, volume~35, pages 12719--12727.

\bibitem[{Fomicheva et~al.(2020{\natexlab{a}})Fomicheva, Sun, Yankovskaya, Blain, Guzm{\'a}n, Fishel, Aletras, Chaudhary, and Specia}]{unsupervised_qe}
Marina Fomicheva, Shuo Sun, Lisa Yankovskaya, Fr{\'e}d{\'e}ric Blain, Francisco Guzm{\'a}n, Mark Fishel, Nikolaos Aletras, Vishrav Chaudhary, and Lucia Specia. 2020{\natexlab{a}}.
\newblock Unsupervised quality estimation for neural machine translation.
\newblock \emph{Transactions of the Association for Computational Linguistics}, 8:539--555.

\bibitem[{Fomicheva et~al.(2020{\natexlab{b}})Fomicheva, Sun, Yankovskaya, Blain, Guzm{\'a}n, Fishel, Aletras, Chaudhary, and Specia}]{unsupervised}
Marina Fomicheva, Shuo Sun, Lisa Yankovskaya, Fr{\'e}d{\'e}ric Blain, Francisco Guzm{\'a}n, Mark Fishel, Nikolaos Aletras, Vishrav Chaudhary, and Lucia Specia. 2020{\natexlab{b}}.
\newblock \href {https://doi.org/10.1162/tacl_a_00330} {Unsupervised quality estimation for neural machine translation}.
\newblock \emph{Transactions of the Association for Computational Linguistics}, 8:539--555.

\bibitem[{Freitag et~al.(2021)Freitag, Foster, Grangier, Ratnakar, Tan, and Macherey}]{MQM}
Markus Freitag, George Foster, David Grangier, Viresh Ratnakar, Qijun Tan, and Wolfgang Macherey. 2021.
\newblock Experts, errors, and context: A large-scale study of human evaluation for machine translation.
\newblock \emph{Transactions of the Association for Computational Linguistics}, 9:1460--1474.

\bibitem[{Gal and Ghahramani(2016)}]{dropout}
Yarin Gal and Zoubin Ghahramani. 2016.
\newblock Dropout as a bayesian approximation: Representing model uncertainty in deep learning.
\newblock In \emph{international conference on machine learning}, pages 1050--1059. PMLR.

\bibitem[{Geng et~al.(2023{\natexlab{a}})Geng, Lai, Zhang, Tao, Yang, Chen, and Huang}]{wmt2023}
Xiang Geng, Zhejian Lai, Yu~Zhang, Shimin Tao, Hao Yang, Jiajun Chen, and Shujian Huang. 2023{\natexlab{a}}.
\newblock Unify word-level and span-level tasks: Njunlp’s participation for the wmt2023 quality estimation shared task.
\newblock In \emph{Proceedings of the Eighth Conference on Machine Translation}, pages 829--834.

\bibitem[{Geng et~al.(2023{\natexlab{b}})Geng, Zhang, Lai, She, Zou, Tao, Yang, Chen, and Huang}]{cbsqe}
Xiang Geng, Yu~Zhang, Zhejian Lai, Shuaijie She, Wei Zou, Shimin Tao, Hao Yang, Jiajun Chen, and Shujian Huang. 2023{\natexlab{b}}.
\newblock \href {https://doi.org/10.18653/v1/2023.emnlp-main.764} {Improved pseudo data for machine translation quality estimation with constrained beam search}.
\newblock In \emph{Proceedings of the 2023 Conference on Empirical Methods in Natural Language Processing}, pages 12434--12447, Singapore. Association for Computational Linguistics.

\bibitem[{Glava{\v{s}} and Vuli{\'c}(2021)}]{glavavs2021climbing}
Goran Glava{\v{s}} and Ivan Vuli{\'c}. 2021.
\newblock Climbing the tower of treebanks: Improving low-resource dependency parsing via hierarchical source selection.
\newblock In \emph{Findings of the Association for Computational Linguistics: ACL-IJCNLP 2021}, pages 4878--4888.

\bibitem[{Graham and Baldwin(2014)}]{williams_test}
Yvette Graham and Timothy Baldwin. 2014.
\newblock Testing for significance of increased correlation with human judgment.
\newblock In \emph{Proceedings of the 2014 Conference on Empirical Methods in Natural Language Processing (EMNLP)}, pages 172--176.

\bibitem[{Graham et~al.(2017)Graham, Baldwin, Moffat, and Zobel}]{DA}
Yvette Graham, Timothy Baldwin, Alistair Moffat, and Justin Zobel. 2017.
\newblock Can machine translation systems be evaluated by the crowd alone.
\newblock \emph{Natural Language Engineering}, 23(1):3--30.

\bibitem[{He et~al.(2024)He, Wang, Jiao, Zhang, Wang, Shi, and Tu}]{QE-as-RM}
Zhiwei He, Xing Wang, Wenxiang Jiao, Zhuosheng Zhang, Rui Wang, Shuming Shi, and Zhaopeng Tu. 2024.
\newblock \href {https://doi.org/10.18653/v1/2024.naacl-long.451} {Improving machine translation with human feedback: An exploration of quality estimation as a reward model}.
\newblock In \emph{Proceedings of the 2024 Conference of the North American Chapter of the Association for Computational Linguistics: Human Language Technologies (Volume 1: Long Papers)}, pages 8164--8180, Mexico City, Mexico. Association for Computational Linguistics.

\bibitem[{Kocmi et~al.(2023)Kocmi, Avramidis, Bawden, Bojar, Dvorkovich, Federmann, Fishel, Freitag, Gowda, Grundkiewicz, Haddow, Koehn, Marie, Monz, Morishita, Murray, Nagata, Nakazawa, Popel, Popovi{\'c}, and Shmatova}]{wmt-2023-mt-findings}
Tom Kocmi, Eleftherios Avramidis, Rachel Bawden, Ond{\v{r}}ej Bojar, Anton Dvorkovich, Christian Federmann, Mark Fishel, Markus Freitag, Thamme Gowda, Roman Grundkiewicz, Barry Haddow, Philipp Koehn, Benjamin Marie, Christof Monz, Makoto Morishita, Kenton Murray, Makoto Nagata, Toshiaki Nakazawa, Martin Popel, Maja Popovi{\'c}, and Mariya Shmatova. 2023.
\newblock \href {https://doi.org/10.18653/v1/2023.wmt-1.1} {Findings of the 2023 conference on machine translation ({WMT}23): {LLM}s are here but not quite there yet}.
\newblock In \emph{Proceedings of the Eighth Conference on Machine Translation}, pages 1--42, Singapore. Association for Computational Linguistics.

\bibitem[{Kocmi and Federmann(2023)}]{gemba-mqm}
Tom Kocmi and Christian Federmann. 2023.
\newblock \href {https://doi.org/10.18653/v1/2023.wmt-1.64} {{GEMBA}-{MQM}: Detecting translation quality error spans with {GPT}-4}.
\newblock In \emph{Proceedings of the Eighth Conference on Machine Translation}, pages 768--775, Singapore. Association for Computational Linguistics.

\bibitem[{Lommel et~al.(2014)Lommel, Burchardt, Popovi{\'c}, Harris, Avramidis, and Uszkoreit}]{MQM_old}
Arle Lommel, Aljoscha Burchardt, Maja Popovi{\'c}, Kim Harris, Eleftherios Avramidis, and Hans Uszkoreit. 2014.
\newblock Using a new analytic measure for the annotation and analysis of mt errors on real data.
\newblock In \emph{Proceedings of the 17th Annual conference of the European Association for Machine Translation}, pages 165--172.

\bibitem[{Luo et~al.(2019)Luo, Xu, Zhang, Zhang, Ren, and Sun}]{pkuseg}
Ruixuan Luo, Jingjing Xu, Yi~Zhang, Zhiyuan Zhang, Xuancheng Ren, and Xu~Sun. 2019.
\newblock \href {https://arxiv.org/abs/1906.11455} {Pkuseg: A toolkit for multi-domain chinese word segmentation.}
\newblock \emph{CoRR}, abs/1906.11455.

\bibitem[{Nivre et~al.(2020)Nivre, De~Marneffe, Ginter, Haji{\v{c}}, Manning, Pyysalo, Schuster, Tyers, and Zeman}]{nivre2020universal}
Joakim Nivre, Marie-Catherine De~Marneffe, Filip Ginter, Jan Haji{\v{c}}, Christopher~D Manning, Sampo Pyysalo, Sebastian Schuster, Francis Tyers, and Daniel Zeman. 2020.
\newblock Universal dependencies v2: An evergrowing multilingual treebank collection.
\newblock \emph{arXiv preprint arXiv:2004.10643}.

\bibitem[{OpenAI(2023)}]{gpt4}
OpenAI. 2023.
\newblock Gpt-4 technical report.
\newblock \emph{arXiv preprint arXiv:2303.08774}.

\bibitem[{Ott et~al.(2019)Ott, Edunov, Baevski, Fan, Gross, Ng, Grangier, and Auli}]{fairseq}
Myle Ott, Sergey Edunov, Alexei Baevski, Angela Fan, Sam Gross, Nathan Ng, David Grangier, and Michael Auli. 2019.
\newblock fairseq: A fast, extensible toolkit for sequence modeling.
\newblock In \emph{Proceedings of NAACL-HLT 2019: Demonstrations}.

\bibitem[{Ouyang et~al.(2022)Ouyang, Wu, Jiang, Almeida, Wainwright, Mishkin, Zhang, Agarwal, Slama, Ray et~al.}]{RLHF}
Long Ouyang, Jeffrey Wu, Xu~Jiang, Diogo Almeida, Carroll Wainwright, Pamela Mishkin, Chong Zhang, Sandhini Agarwal, Katarina Slama, Alex Ray, et~al. 2022.
\newblock Training language models to follow instructions with human feedback.
\newblock \emph{Advances in Neural Information Processing Systems}, 35:27730--27744.

\bibitem[{Papineni et~al.(2002)Papineni, Roukos, Ward, and Zhu}]{BLEU}
Kishore Papineni, Salim Roukos, Todd Ward, and Wei-Jing Zhu. 2002.
\newblock Bleu: a method for automatic evaluation of machine translation.
\newblock In \emph{Proceedings of the 40th annual meeting of the Association for Computational Linguistics}, pages 311--318.

\bibitem[{Qi et~al.(2020)Qi, Zhang, Zhang, Bolton, and Manning}]{qi2020stanzapythonnaturallanguage}
Peng Qi, Yuhao Zhang, Yuhui Zhang, Jason Bolton, and Christopher~D. Manning. 2020.
\newblock \href {https://arxiv.org/abs/2003.07082} {Stanza: A python natural language processing toolkit for many human languages}.
\newblock \emph{Preprint}, arXiv:2003.07082.

\bibitem[{Rei et~al.(2023)Rei, Guerreiro, Pombal, van Stigt, Treviso, Coheur, C.~de Souza, and Martins}]{CometKiwi2023}
Ricardo Rei, Nuno~M. Guerreiro, Jos{\~A}{\copyright} Pombal, Daan van Stigt, Marcos Treviso, Luisa Coheur, Jos{\'e}~G. C.~de Souza, and Andr{\'e} Martins. 2023.
\newblock \href {https://doi.org/10.18653/v1/2023.wmt-1.73} {Scaling up {C}omet{K}iwi: Unbabel-{IST} 2023 submission for the quality estimation shared task}.
\newblock In \emph{Proceedings of the Eighth Conference on Machine Translation}, pages 841--848, Singapore. Association for Computational Linguistics.

\bibitem[{Rei et~al.(2020)Rei, Stewart, Farinha, and Lavie}]{comet}
Ricardo Rei, Craig Stewart, Ana~C Farinha, and Alon Lavie. 2020.
\newblock \href {https://doi.org/10.18653/v1/2020.emnlp-main.213} {{COMET}: A neural framework for {MT} evaluation}.
\newblock In \emph{Proceedings of the 2020 Conference on Empirical Methods in Natural Language Processing (EMNLP)}, pages 2685--2702, Online. Association for Computational Linguistics.

\bibitem[{Scarton et~al.(2015)Scarton, Tan, and Specia}]{scarton2015ushef}
Carolina Scarton, Liling Tan, and Lucia Specia. 2015.
\newblock Ushef and usaar-ushef participation in the wmt15 qe shared task.
\newblock In \emph{Proceedings of the Tenth Workshop on Statistical Machine Translation}, pages 336--341.

\bibitem[{Shah et~al.(2015)Shah, Ng, Bougares, and Specia}]{shah2015investigating}
Kashif Shah, Raymond~WM Ng, Fethi Bougares, and Lucia Specia. 2015.
\newblock Investigating continuous space language models for machine translation quality estimation.
\newblock In \emph{Proceedings of the 2015 Conference on Empirical Methods in Natural Language Processing}, pages 1073--1078.

\bibitem[{Snover et~al.(2006)Snover, Dorr, Schwartz, Micciulla, and Makhoul}]{hter}
Matthew Snover, Bonnie Dorr, Richard Schwartz, Linnea Micciulla, and John Makhoul. 2006.
\newblock A study of translation edit rate with targeted human annotation.
\newblock In \emph{Proceedings of association for machine translation in the Americas}, volume 200. Cambridge, MA.

\bibitem[{Specia(2011)}]{effort}
Lucia Specia. 2011.
\newblock Exploiting objective annotations for minimising translation post-editing effort.
\newblock In \emph{Proceedings of the 15th Annual conference of the European Association for Machine Translation}.

\bibitem[{Specia et~al.(2018)Specia, Scarton, and Paetzold}]{mtqe}
Lucia Specia, Carolina Scarton, and Gustavo~Henrique Paetzold. 2018.
\newblock Quality estimation for machine translation.
\newblock \emph{Synthesis Lectures on Human Language Technologies}, 11(1):1--162.

\bibitem[{Specia et~al.(2013)Specia, Shah, De~Souza, and Cohn}]{quest}
Lucia Specia, Kashif Shah, Jos{\'e}~GC De~Souza, and Trevor Cohn. 2013.
\newblock Quest-a translation quality estimation framework.
\newblock In \emph{Proceedings of the 51st Annual Meeting of the Association for Computational Linguistics: System Demonstrations}, pages 79--84.

\bibitem[{Tuan et~al.(2021)Tuan, El-Kishky, Renduchintala, Chaudhary, Guzm{\'a}n, and Specia}]{nmt_ter}
Yi-Lin Tuan, Ahmed El-Kishky, Adithya Renduchintala, Vishrav Chaudhary, Francisco Guzm{\'a}n, and Lucia Specia. 2021.
\newblock \href {https://doi.org/10.18653/v1/2021.eacl-main.50} {Quality estimation without human-labeled data}.
\newblock In \emph{Proceedings of the 16th Conference of the European Chapter of the Association for Computational Linguistics: Main Volume}, pages 619--625, Online. Association for Computational Linguistics.

\bibitem[{Vaswani et~al.(2017)Vaswani, Shazeer, Parmar, Uszkoreit, Jones, Gomez, Kaiser, and Polosukhin}]{transformer}
Ashish Vaswani, Noam Shazeer, Niki Parmar, Jakob Uszkoreit, Llion Jones, Aidan~N. Gomez, Lukasz Kaiser, and Illia Polosukhin. 2017.
\newblock Attention is all you need.
\newblock \emph{Advances in Neural Information Processing Systems 30}.

\bibitem[{Xu et~al.(2024{\natexlab{a}})Xu, Deutsch, Finkelstein, Juraska, Zhang, Liu, Wang, Li, and Freitag}]{llmrefine}
Wenda Xu, Daniel Deutsch, Mara Finkelstein, Juraj Juraska, Biao Zhang, Zhongtao Liu, William~Yang Wang, Lei Li, and Markus Freitag. 2024{\natexlab{a}}.
\newblock \href {https://doi.org/10.18653/v1/2024.findings-naacl.92} {{LLMR}efine: Pinpointing and refining large language models via fine-grained actionable feedback}.
\newblock In \emph{Findings of the Association for Computational Linguistics: NAACL 2024}, pages 1429--1445, Mexico City, Mexico. Association for Computational Linguistics.

\bibitem[{Xu et~al.(2024{\natexlab{b}})Xu, Li, Wang, and Li}]{bpo}
Wenda Xu, Jiachen Li, William~Yang Wang, and Lei Li. 2024{\natexlab{b}}.
\newblock \href {https://aclanthology.org/2024.emnlp-main.623} {{BPO}: Staying close to the behavior {LLM} creates better online {LLM} alignment}.
\newblock In \emph{Proceedings of the 2024 Conference on Empirical Methods in Natural Language Processing}, pages 11125--11139, Miami, Florida, USA. Association for Computational Linguistics.

\bibitem[{Xu et~al.(2023)Xu, Wang, Pan, Song, Freitag, Wang, and Li}]{instructscore}
Wenda Xu, Danqing Wang, Liangming Pan, Zhenqiao Song, Markus Freitag, William Wang, and Lei Li. 2023.
\newblock \href {https://doi.org/10.18653/v1/2023.emnlp-main.365} {{INSTRUCTSCORE}: Towards explainable text generation evaluation with automatic feedback}.
\newblock In \emph{Proceedings of the 2023 Conference on Empirical Methods in Natural Language Processing}, pages 5967--5994, Singapore. Association for Computational Linguistics.

\bibitem[{Zheng et~al.(2021)Zheng, Tan, Zhang, Maimaiti, Luan, Sun, Liu, and Liu}]{ssqe}
Yuanhang Zheng, Zhixing Tan, Meng Zhang, Mieradilijiang Maimaiti, Huanbo Luan, Maosong Sun, Qun Liu, and Yang Liu. 2021.
\newblock \href {https://doi.org/10.18653/v1/2021.emnlp-main.267} {Self-supervised quality estimation for machine translation}.
\newblock In \emph{Proceedings of the 2021 Conference on Empirical Methods in Natural Language Processing}, pages 3322--3334, Online and Punta Cana, Dominican Republic. Association for Computational Linguistics.

\bibitem[{Zouhar et~al.(2024)Zouhar, Chen, Lam, Moghe, and Haddow}]{zouhar2024pitfalls}
Vil{\'e}m Zouhar, Pinzhen Chen, Tsz~Kin Lam, Nikita Moghe, and Barry Haddow. 2024.
\newblock Pitfalls and outlooks in using comet.
\newblock In \emph{Proceedings of the Ninth Conference on Machine Translation}, pages 1272--1288.

\end{thebibliography}

\appendix

\section{Experiment Details}
\label{sec:appendix oid}

\subsection{Data statistics}
\label{sec:data_statics}
\begin{table}[h] 
\centering
\resizebox{0.48\textwidth}{!}{
\begin{tabular}{cccc}
\toprule
\textbf{Type} & \textbf{Dataset} & \textbf{Split} & \textbf{Size} \\
\midrule
\multirow{3}{*}{Parallel data} & WMT23 EN-DE & Train & 45M \\
& WMT23 ZH-EN & Train & 30M \\
& WMT23 HE-EN & Train & 35M \\
\midrule
\multirow{7}{*}{MQM data} & WMT23 EN-DE & Train \& Valid & 150K \\
& WMT23 ZH-EN & Train \& Valid & 200K \\
& WMT23 EN-DE & Test & 1887 \\
& WMT23 ZH-EN & Test & 1664 \\
& WMT23 HE-EN & Test & 1134 \\
& WMT22 EN-DE & Test & 511 \\
& WMT22 ZH-EN & Test & 505 \\
\bottomrule
\end{tabular}}
\caption{Statistics of the datasets.}
\label{tab:dataset_info}
\end{table}

\subsection{Generator \& Annotator}
The model parameters are optimized using the Adam optimizer, configured with  $\beta_1 = 0.9$ and $\beta_2 = 0.98$.
The initial learning rate is set to 5e-4, and the inverse square root learning rate scheduler is employed with 6000 warm-up steps.
Model parameters update every 20 batches and the batch is configured to  13,650.
The dropout rate is set to 0.3, and the weight decay is set to 1e-4.
The training objective employs the label-smoothed cross-entropy criterion with $\text{eps}=0.1$. 
The training stops early if there are no improvements in validation performance for the last 15 epochs.

We use the TER tool called TERCOM\footnote{\url{http://www.cs.umd.edu/~snover/tercom/}} to annotate the synthetic translations generated by the Generator.

\subsection{InstructScore}
By providing the references only, Instructscore prompts the GPT-4 to generate German and English synthetic translations along with explainable texts. 
We get these data from their public repository\footnote{\url{https://github.com/xu1998hz/InstructScore_SEScore3}}.
To construct the synthetic sources, we translated the German sentences into English and translated the English sentences into Chinese and Hebrew using Google Translate.
Following this, regular expressions are employed to extract MQM labels from the explainable texts.
These labels are then systematically organized alongside the corresponding sources and translations, resulting in a unified MQM dataset structured for analysis.

\subsection{GEMBA-MQM}
We adopt the same configuration as GEMBA-MQM~\cite{gemba-mqm}, but utilize the \textit{gpt-4-0125-preview} model, and structured QE data is extracted from the regular expression.

\subsection{Preprocess}
We use the sacremoses toolkit\footnote{\url{https://github.com/alvations/sacremoses}} to normalize and tokenize the parallel sentences. For the Chinese sentences, we utilize pkuseg~\cite{pkuseg} for Chinese word segmentation.

\subsection{Pretraining and Finetuning}
For unsupervised experiments, 4 NVIDIA V100 GPUs are used to train the QE models. 
The learning rate is set to 1e-6 for the EN-DE direction and 1e-5 for the ZH-EN \& HE-EN direction.
The model parameters are optimized using the Adam optimizer, configured with  $\beta_1 = 0.9$ and $\beta_2 = 0.999$, and the clip norm is set to 1.0
During training, we set the maximum number of sentences in a batch to 15, and the update frequency is set to 20. 
The word-level ``OK'' label weight is twice the ratio of ``BAD'' to ``OK'' labels in the synthetic data while the ``BAD'' label weight is fixed to 2.0.
The training stops early if there are no improvements in validation performance for the last 15 epochs.

For supervised experiments, one NVIDIA V100 GPU is used to train the models.
The model parameters are optimized using the Adam optimizer, configured with  $\beta_1 = 0.9$ and $\beta_2 = 0.999$, and the clip norm is set to 1.0. 
The learning rate is set to 1e-6 for the EN-DE direction and 7e-6 for ZH-EN direction.
During training, we set the maximum number of sentences in a batch to 15, and the update frequency is set to 20. 
The training stops early if there are no improvements in validation performance for the last 15 epochs.

\subsection{Inference}

For the span-level task, we performed a greedy search to optimize thresholds for categorizing MINOR, MAJOR, and CRITICAL severity for each language direction. This optimization is conducted on the WMT2023 QE validation set for EN-DE and ZH-EN, and on a subset of the test set with 100 entries for HE-EN.
The t
For the sentence-level task, we calculated scores by averaging the regression score and the MQM score derived from the span-level results.

\section{Details of SPCE algorithm}
\label{sec:spce_details}
The pseudo-code for the implementation of the SPCE algorithm is presented in Algorithm~\ref{alg:qe}.

The SPCE algorithm depends on models for dependency parsing. However, compared to QE tasks, dependency parsing is a well-established NLP task that has been extensively studied for decades, with significantly more available resources. 
For instance, the Universal Dependencies (UD) project~\cite{nivre2020universal} covers over 150 languages with 200 treebanks. 
To further imporve the performance on low-resource languages, TowerParse~\cite{glavavs2021climbing} leverages a heuristic data selection strategy to automatically identify suitable training data from UD for any target language, achieving competitive results.
In this work, we utilize the Stanza toolkit~\cite{qi2020stanzapythonnaturallanguage} to perform dependency parsing on the synthetic translations. 

\begin{algorithm*}[h]
    \caption{Shortest Phrase Covering Errors.}
    \label{alg:qe}
    \textbf{Input}: {The error interval $\{l, \dots, r\}$, the dependency tree $\mathcal{T}$}\\
    \textbf{Output}: {The shortest phrase covering errors $\{\hat l, \dots,  \hat r\}$.}
    \begin{algorithmic}[1]
    \STATE $\mathcal{P}_{cur}\longleftarrow\{l, \dots, r\}$,  $\mathcal{P}_{last}\longleftarrow\emptyset$
    \WHILE{$\mathcal{P}_{last}\neq \mathcal{P}_{cur}$}
    \STATE $\mathcal{P}_{last} \longleftarrow \mathcal{P}_{cur}$\; 
    \STATE \# Least common ancestor of selected words in the dependency tree $\mathcal{T}$.
    \STATE $a := \text{LCA}(\mathcal{T},\mathcal{P}_{cur})$\; 
    \STATE \# Make the selected words form a dependency subtree.
    \FORALL{$p\in\mathcal{P}_{cur}$}
        \WHILE{$p$ is not $a$} 
                \STATE $p :=\text{get\_parent}(\mathcal{T}, p)$\;
                \STATE $\mathcal{P}_{cur} \longleftarrow \mathcal{P}_{cur} \cup \{ p \}$\; 
        \ENDWHILE
    \ENDFOR
    \STATE \# Make the selected words that make up the phrase consecutive.
    \STATE $\hat{l}:=\min(\mathcal{P}_{cur})$, $\hat{r}:=\max(\mathcal{P}_{cur})$
    \FORALL{$i \in \{\hat l, \dots,  \hat r\}$} %
        \STATE $\mathcal{P}_{cur} \longleftarrow \mathcal{P}_{cur} \cup \{i\}$\;  
    \ENDFOR\;
    \ENDWHILE
    \RETURN $\{\hat l, \dots,  \hat r\}$
    \end{algorithmic}
\end{algorithm*}

\section{Case Study}

Section~\ref{sec:annotate_itself} demonstrates that the model exhibits overconfidence in its own outputs, resulting in erroneous ``OK'' labels. For example, in Table~\ref{tab:annotate_itself_case_study}, the words ``sind'' and ``einem'' are incorrectly labeled as ``OK'' by model $M$ itself, despite being grammatical errors. Deploying another model $M'$ as the Annotator helps mitigate this issue. 
Leveraging the SPCE algorithm, the phrase ``sind in einem'' is correctly identified as a complete unit and labeled as ``BAD''. "privaten Vorstellung" and ``Privatshow'' are synonyms, and the former has been correctly rejudged as ``OK''.

\begin{table}[h]
\centering
\resizebox{\columnwidth}{!}{
\begin{tabular}{l|l}
\hline
\textbf{Source} &	Helena Anderson is in a private show \\
\textbf{Reference} &	Helena Anderson sich in einer Privatshow \\
\hline
\hline
\textbf{Generator $M$} &	Helena Anderson sind in einem privaten Vorstellung \\
\textbf{Annotator $M$} &	OK OK OK OK OK OK OK \\
\textbf{Annotator $M'$} &	OK OK \textcolor{red}{BAD} \textcolor{red}{BAD} \textcolor{red}{BAD} OK OK \\
\hline
\end{tabular}
}
\caption{A Case Study illustrating the model exhibits overconfidence in its own outputs.
Model $M$ consistently assigning the OK labels to all tokens generated by itself.}
\label{tab:annotate_itself_case_study}
\end{table}

Section~\ref{sec:enhance_annotator} demonstrates that supervision signals is helpful for enhancing the capacity of annotation model.
In Table~\ref{tab:annotator_case_study}, the word ``Anträge'' is correctly labeled as ``BAD'' by model $M_\text{amateur}$. However, the assigned error severity is incorrect. Since Annotator $M'$ has been trained on this specific parallel sentence, it provides the correct error severity and additionally identifies another error, the word ``aufgenommen''. Furthermore, Annotator $L'$, possessing more advanced translation capability, produces entirely accurate annotations.

\begin{table}[h]
\centering
\resizebox{\columnwidth}{!}{
\begin{tabular}{l|l|l}
\hline
\textbf{Source} & \multicolumn{2}{l}{Several anecdotes are included .} \\
\textbf{Reference} & \multicolumn{2}{l}{Viele Anekdoten sind darüber überliefert .} \\
\hline
\hline
\textbf{Generator $M$} & \multicolumn{2}{l}{Es werden mehrere Anträge aufgenommen .} \\
\textbf{Annotator $M_\text{amateur}$} & \multicolumn{2}{l}{OK OK OK \textcolor{red}{MAJOR} OK OK}\\
\textbf{Annotator $M'$} & \multicolumn{2}{l}{OK OK OK \textcolor{red}{CRITICAL} \textcolor{red}{CRITICAL} OK}\\
\textbf{Annotator $L'$} & \multicolumn{2}{l}{OK OK OK \textcolor{red}{CRITICAL} \textcolor{red}{MINOR} OK}\\
\hline
\end{tabular}
}
\caption{A Case Study illustrating the distinction between Annotator $M_\text{amateur}$, $M'$ and $L$ with Generator $M$.}
\label{tab:annotator_case_study}
\end{table}

\begin{table}[htbp]
\centering
\resizebox{\columnwidth}{!}{\begin{tabular}{cccc}
\toprule
 \textbf{Slabele} & \textbf{Device} & \textbf{Calculation Time (ms)} & \textbf{Proportion (\%)} \\
\midrule
1 & GPU & 10.37 & 10.13 \\
2 & CPU & \hspace{5pt}0.24 & \hspace{5pt}0.23 \\
3 & GPU & \hspace{5pt}4.11 & \hspace{5pt}4.02 \\
4 & CPU \& GPU & 87.64 & 85.62 \\
\bottomrule
\end{tabular}}
\caption{Time Analysis of DCSQE.}
\label{tab:time_analysis}
\end{table}

\begin{table*}[h!]
\centering
\begin{tabular}{lcc}
\toprule
\textbf{Translation Error Type} & \textbf{w/o Synthetic Data} & \textbf{w/ Synthetic Data} \\
\midrule
accuracy/addition & 74.22\% & 85.39\% \\
accuracy/creative reinterpretation & 21.42\% & 35.57\% \\
accuracy/mistranslation & 43.50\% & 59.23\% \\
accuracy/source language fragment & 46.09\% & 55.87\% \\
fluency/grammar & 31.31\% & 47.79\% \\
fluency/inconsistency & 26.60\% & 38.38\% \\
fluency/punctuation & 32.64\% & 42.39\% \\
fluency/register & 27.00\% & 39.00\% \\
fluency/spelling & 32.91\% & 43.18\% \\
locale convention/currency format & 56.67\% & 56.67\% \\
locale convention/date format & 50.00\% & 100.00\%\ \ \  \\
locale convention/time format & \ 
\ 0.00\% & 40.00\% \\
style/bad sentence structure & 31.95\% & 38.96\% \\
style/unnatural or awkward & 32.01\% & 43.47\% \\
terminology/inappropriate for context & 27.74\% & 38.41\% \\
terminology/inconsistent & 26.67\% & 33.33\% \\
non-translation! & 26.38\% & 63.83\% \\
other & 43.07\% & 53.92\% \\
\bottomrule
\end{tabular}
\caption{Comparison of error detection accuracy with and without synthetic data across translation error types.}
\label{tab:error_type_improvement}
\end{table*}

\section{Time Analysis}
\label{sec:time_analysis}

DCSQE can be divided into four stages: (1) Synthesize Translations with the Generator, (2) Synthesize coarse-grained labels with TER tool, (3) Refining labels with the Annotator, (4) Aggregate the results with the SPCE algorithm.
We counted the time spent on each stage in Table~\ref{tab:time_analysis}.

\section{Convergence Efficiency.}
\begin{figure}[h]
    \centering
    \includegraphics[width=\columnwidth]{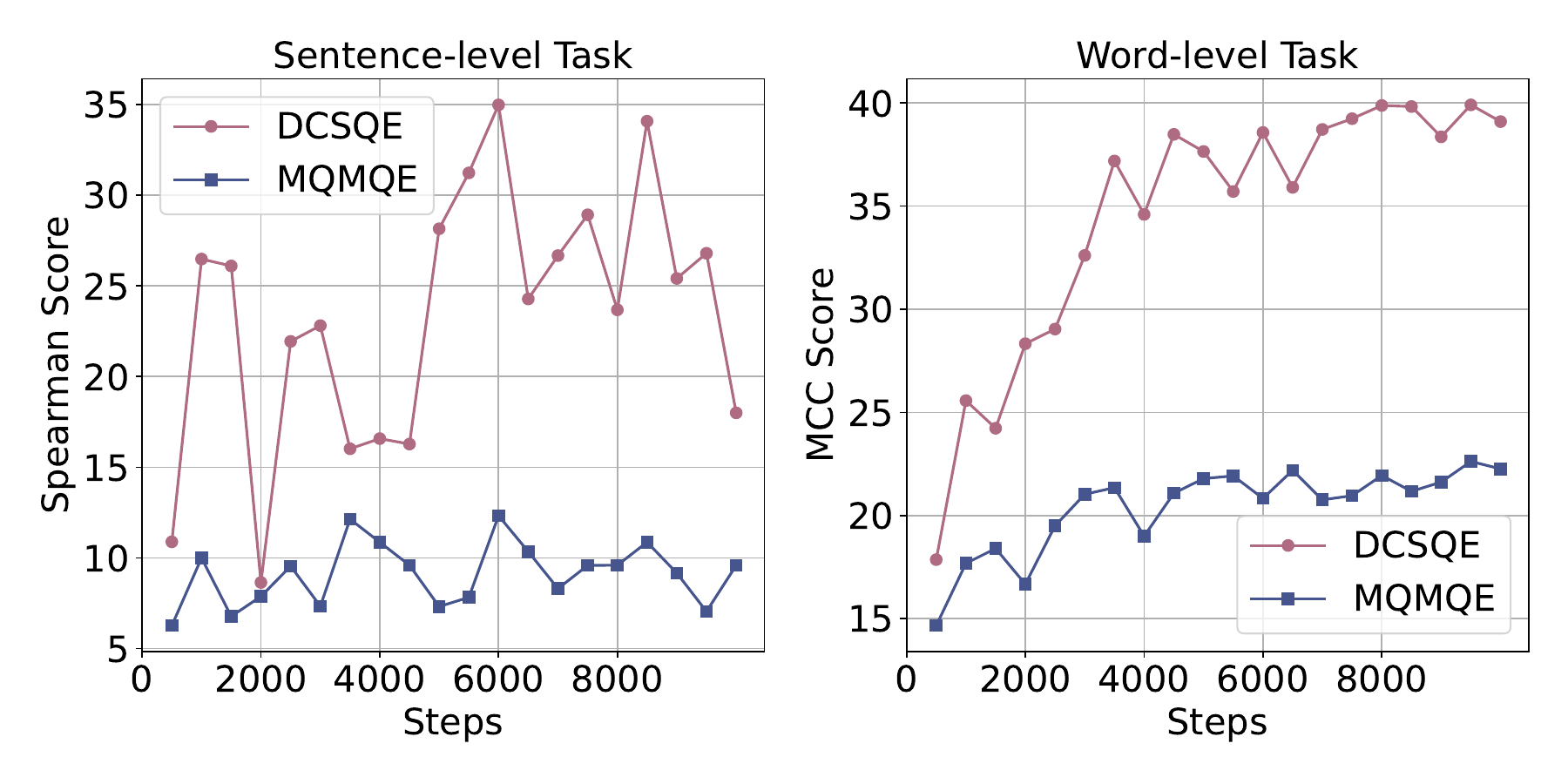}
    \caption{Training steps vs. Spearman or MCC score of different methods on the different task in WMT23 EN-DE QE validation set.}
    \label{fig:generation_convergence}
\end{figure}

Higher-quality synthetic data can accelerate training convergence. 
To evaluate the convergence efficiency of different methods, we plot the learning curve on the WMT2022 validation set in Figure~\ref{fig:generation_convergence}.
For both sentence-level and word-level tasks, CBSQE demonstrates faster convergence compared to MQMQE, ultimately achieving the highest performance.

\section{Amount of Synthetic Data}

\begin{figure}[h]
    \centering
    \includegraphics[width=\columnwidth]{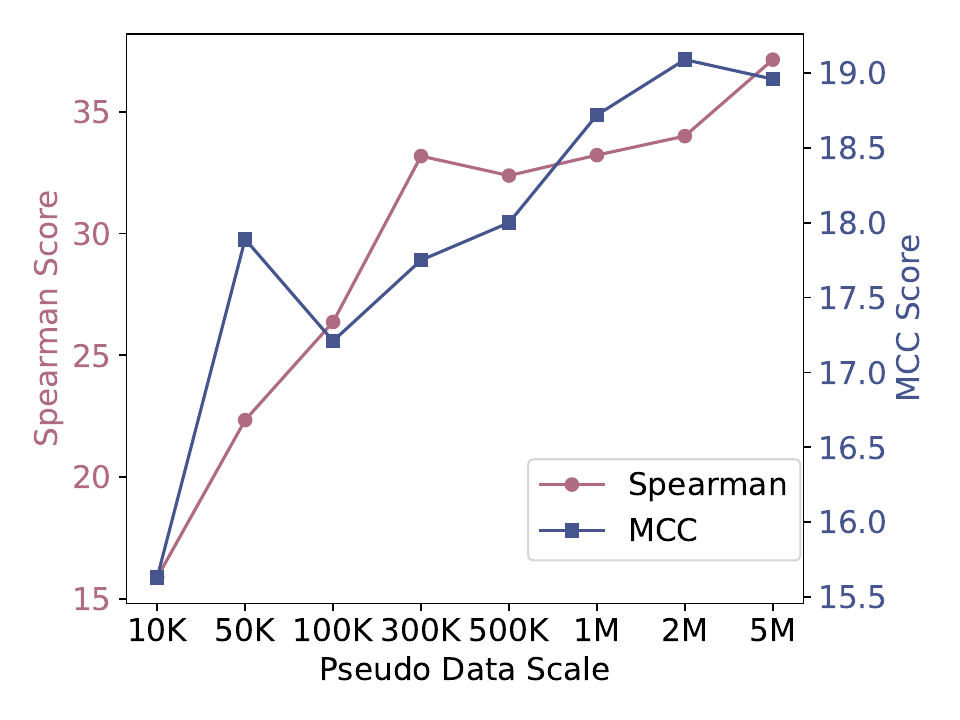}
    \caption{Spearson or MCC score on the WMT23 EN-DE QE test set using different amounts of parallel pairs.}
    \label{fig:data_scale}
\end{figure}

We examine the impact of data size on the performance of DCSQE.
To this end, we conduct a series of experiments utilizing varying amounts of parallel sentence pairs from the WMT23 EN-DE dataset.
As illustrated in Figure~\ref{fig:data_scale}, for all the tasks, synthetic data methods demonstrate significant performance improvements as the data size increases.
Notably, this upward trend persists and does not exhibit convergence until reaching a scale of 5M parallel pairs.

\section{Threshold Sensitivity}
In DCSQE, the thresholds can be easily acquired by small size validation sets for each language pair which are available in most real world scenarios.
For a given language pair, these thresholds exhibit robustness across different datasets. 

In our experiments, we apply the same thresholds derived from the WMT2022 validation dataset to both the WMT2022 and WMT2023 test sets. 
As shown in Table~\ref{tab:main_result}, DCSQE still achieves consistent performance across different test sets for both EN-DE and ZH-EN language pairs. 
Notably, there exist significant differences between WMT2022 and WMT2023 test sets as discussed in~\cite{wmt2023-findings}.

\section{Completeness of Error Modeling}

Alignment between translations and references is generally effective in identifying most translation errors, as these typically appear as discrepancies with the references. 
Our manual inspection of 100 randomly sampled synthetic instances, does not reveal any undetected errors (i.e., false-positives labels). 
This observation is consistent with the finding in CBSQE~\cite{cbsqe}, which reports that the TER tool frequently produces false-negative labels but rarely false positives.

To more comprehensively validate the comprehensiveness of error modeling in DCSQE, we compare the error detection accuracy of models without/with DCSQE synthetic data pre-training across various error categories. Since error category labels are only available in training data, we use the WMT2024 training set\footnote{\url{https://github.com/google/wmt-mqm-human-evaluation/tree/main/generalMT2023}} which is never exposed during our model training. The experimental results demonstrate significant improvements in nearly all error categories (including semantic errors), as shown in the Table~\ref{tab:error_type_improvement}. This finding confirms that our synthetic data encompasses diverse error types.

\end{document}